\relax
\documentclass[letterpaper]{article} 

\usepackage{aaai19}  
\usepackage{times}  
\usepackage{helvet}  
\usepackage{courier}  
\usepackage{url}  
\usepackage{graphicx}  
\usepackage{color}
\usepackage{caption}
\usepackage{booktabs}
\usepackage{threeparttable}
\usepackage{multirow}
\usepackage{makecell}
\usepackage{bm}
\usepackage{hyperref}       
\usepackage{url}            

\usepackage{helvet}
\usepackage{courier}
\usepackage{graphicx}
\usepackage{booktabs}
\usepackage{amsmath}
\usepackage{amssymb}
\usepackage{subfigure}
\usepackage{algorithm}
\usepackage{algorithmic}

\def\b{{\bf b}}

\def\h{{\bf h}}

\def\x{{\bf x}}

\def\U{{\bf U}}

\def\W{{\bf W}}

\def\0{{\bf 0}}
\def\1{{\bf 1}}

\begin{document}
%

\title{Recurrent Attention Unit}
\author{Guoqiang Zhong, Guohua Yue and Xiao Ling\\
Department of Computer Science and Technology, Ocean University of China,\\
238 Songling Road, Qingdao, China 266100.\\
Email: gqzhong@ouc.edu.cn, 15053241153@163.com, mumubuguai@vip.qq.com.\\
}

\maketitle
\begin{abstract}
Recurrent Neural Network (RNN) has been successfully applied in many sequence learning problems. Such as handwriting recognition, image description, natural language processing and video motion analysis. After years of development, researchers have improved the internal structure of the RNN and introduced many variants. Among others, Gated Recurrent Unit (GRU) is one of the most widely used RNN model. However, GRU lacks the capability of adaptively paying attention to certain regions or locations, so that it may cause information redundancy or loss during leaning. In this paper, we propose a RNN model, called Recurrent Attention Unit (RAU), which seamlessly integrates the attention mechanism into the interior of GRU by adding an attention gate. The attention gate can enhance GRU's ability to remember long-term memory and help memory cells quickly discard unimportant content. RAU is capable of extracting information from the sequential data by adaptively selecting a sequence of regions or locations and pay more attention to the selected regions during learning. Extensive experiments on image classification, sentiment classification and language modeling show that RAU consistently outperforms GRU and other baseline methods.

\end{abstract}

\section{Introduction}
Recurrent Neural Network (RNN) is a popular method in the field of deep learning in recent years, and it has been proved to be very successful in both classification and generation tasks (see, e.g.\cite{Graves13,Bahdanau2014Neural,SutskeverVL14}). Unlike other artificial neural networks, such as deep belief nets \cite{Hinton:2006:FLA:1161603.1161605} and convolutional neural networks \cite{L1998Gradient}, RNN is commonly used to deal with sequence modeling problems. One of the core ideas of RNN is to connect previous information with the current processing unit, that is, to have a memory function for the previous information. Theoretically RNN can capture any long-term dependency information. However, it is difficult to do so in practical applications due to the gradient exploding and the vanishing problems \cite{hochreiter91,BengioSF94,Hochreiter98}. One of the most promising ways to solve this problem is to change the architecture of the RNN, e.g., by using a gated activation function to achieve the trade-off between the old time information and the new time information. The essence is to adjust the memory focus according to the training target and then perform the whole string coding. Both long short-term memory (LSTM) \cite{HochreiterS97} and gated recurrent unit (GRU) \cite{ChoMGBBSB14} architectures use gated activation functions, which allow the network to learn long-term dependency information and alleviate the gradient vanishing and exploding problems.

Both GRU and LSTM are extensions of the RNN model, but compared to LSTM, GRU reduces the number of gate control units from 3 to 2, and the model is simpler and has higher efficiency. \cite{JozefowiczZS15} compared GRU with LSTM, finding that GRU can achieve results equivalent to LSTM on many issues and is easier to train. Therefore, GRU are increasingly used in natural language processing tasks. For example, \cite{ShangLL15} used GRU to their neural response machines, and \cite{KirosZSZUTF15} also replaced the traditional LSTM with GRU in their implementation of the language model. In this work, we focus on GRU because they have been shown easier to train than LSTM and have good performance.

Attention mechanism is a resource allocation mechanism. For example, in human visual processing, even if human eyes have the ability to accept a large visual field, people's eyes are usually fixed on a certain part. That is, with limited visual resources, one needs to firstly select a specific part of the visual field, and then focus on it. For example, when one is reading, the word to be read is usually processed at a specific time. Just like people's perception mechanisms, computer vision should also focus on a specific part of the input, rather than assigning all inputs the same weight. Therefore, the attention mechanism is a simulation of human attention allocation mechanism, and is used to allocate limited resources to important parts. The nature of attention has been studied extensively in the previous literatures \cite{WaltherIRPK02,itti2001,MnihHGK14,ZhaoWFPY17,DBLP:journals/corr/WangT16}. The attention mechanism is originally applied only to image recognition tasks in computer vision and is subsequently applied to graphic conversions. In natural language processing, the attention mechanism is usually used combining with the Encoder-Decoder model, and the application scenario is very extensive.

We have found that although GRU solves the problem of memorizing longer-term information, it does not highlight the important content of the information. Fortunately, the learning model based on the attention mechanism can pay close attention to the part that needs attention and suppresses unimportant information. Motivated by it,  we propose the Recurrent Attention Unit (RAU),  a novel RNN  architecture that combines the strengths of both GRU and attention mechanism. In RAU, we apply the attention model to the interior of GRU, by seamlessly adding an attention gate. The attention gate enhances GRU's ability to remember long-term memory and help memory cells quickly discard unimportant content.

In this paper, we make the following contributions:

1. We propose a novel RNN architecture called RAU. Our experiments demonstrate that our RAU significantly outperforms than LSTM and GRU models.

2. Usually, attention mechanisms are implemented as additional layer connected to the original RNN. Both the original RNN and the attention module should be learned simultaneously. On contrast, RAU seamlessly adds the attention gate in the memory cell of GRU, which makes the model simpler and easier to train.

3. Not limited in the field of computer vision, our proposed RAU model can also be applied to all sequence related problems. Our experiments show that RAU is suitable for image classification, language modeling and sentiment classification tasks.

\section{Related Work}
 Theoretically, Recurrent Neural Network (RNN) can capture any long-term dependence of the input sequence. But in practice, RNNs must maintain an activation vector for each time step, which makes the RNN very deep. And this depth, in turn, makes RNN suffer from  gradient vanishing and exploding problems \cite{BengioSF94}, making the RNN difficult to train. There have been a number of attempts to address the problem of training RNN. Long Short-Term Memory (LSTM) is an improved RNN model proposed by \cite{HochreiterS97}. It can effectively alleviate the long-distance dependence problem of RNN by constructing special memory units to store historical information, so that each time state preserves the previous input information. Later, \cite{ChoMGBBSB14} proposed a simpler and more convenient Gated Recurrent Unit (GRU) architecture, which not only has the advantage of LSTM, but also is easier to train and has good performance. \cite{DBLP:journals/corr/BradburyMXS16} proposed a new network architecture that replaces the traditional loop structure (vanilla RNN, LSTM, GRU) with a convolution operation. QRNN can be thought as a special structure between RNN and CNN. Since the convolution operation has no time dependence on the loop structure, the calculation parallelism of the QRNN is high. There are also many studies that use the hidden layer unit of LSTM to construct a multi-dimensional \cite{Graves2007Multi} or grid-structured \cite{DBLP:journals/corr/KalchbrennerDG15} LSTM.

 The importance of the attention mechanism can be seen from the human perception process, which uses top information to guide the bottom-up feedback process. Recently, many studies have focused on applying attention mechanisms to deep neural networks. The attention mechanism was first proposed in the field of visual imagery. Subsequently, the Google DeepMind team used the attention mechanism on the RNN model to classify images and achieve good results, which proposed in the \cite{DBLP:journals/corr/MnihHGK14}. \cite{Bahdanau2014Neural} used an attention-like mechanism to translate and align on machine translation tasks simultaneously, this work is the first to propose an attention mechanism to be applied to the natural language processing (NLP) domain. Then a similar extension mechanism based on RNN was applied to various NLP tasks. Recently, how to use the attention mechanism in deep neural network has become a hot research topic in the deep learning area.

 There are many classifications of attention mechanism. \cite{XuBKCCSZB15} introduced attention in image caption. \cite{XuBKCCSZB15} used two attention mechanisms, soft attention and hard attention. Soft attention is parameterized, so it can be guided and embedded in the model for direct training. Gradient can be passed back through the attention mechanism module to other parts of the model. In contrast, hard attention is a random process, which does not select the output of the entire encoder as its input. Hard attention samples the hidden state of the input according to the probability, instead of the hidden state of the entire encoder. In order to achieve backpropagation of the gradient, Monte Carlo sampling is needed to estimate the gradient of the module. Both attention mechanisms have their own advantages, but more researches and applications are now inclined more to use soft attention because it can be directly derived and gradient backpropagation. \cite{DBLP:journals/corr/LuongPM15} proposed two kinds of attention mechanisms: the global attention mechanism and the local attention mechanism. The difference is whether one concerns with all encoder states or partial encoder states. \cite{DBLP:journals/corr/VaswaniSPUJGKP17} proposed self-attention mechanism, which is performed on the source and target sides respectively. Only the self-attention associated with the source input or the target input itself captures the dependency between the source and the target itself. Therefore, self-attention is better than the traditional attention mechanism. One of the main reasons is that the traditional attention mechanism ignores the dependence between words and words in the source or target sentences. In contrast, self-attention can not only get the dependencies between the source and the target, but also effectively get the dependencies between the words of the source or the target.

 In the image classification task, the top-down attention mechanism is widely used. \cite{DBLP:journals/corr/GregorDGW15} modelled image classification as a sequential process, which established a two-dimensional differentiable attention mechanism embedded in DRAW \cite{DBLP:journals/corr/GregorDGW15} and achieved image classification. \cite{XuBKCCSZB15} used the soft attention mechanism to achieve different image areas at different times of decoding, which in turn can generate more reasonable words. An attention map is output in the residual attention network  \cite{DBLP:journals/corr/WangJQYLZWT17}, which enhances meaningful features while suppressing meaningless information. Thus, the attention mechanism can be applied to image classification tasks, and captures different types of attention in a target-driven manner.

 In natural language processing tasks, the google machine translation team proposed a completely attention based architecture \cite{DBLP:journals/corr/VaswaniSPUJGKP17}  which uses a lot of self-attention mechanism to learn text representation. Different from previous machine translation using RNN-based seq2seq model framework, the paper used the attention mechanism instead of RNN to build the entire model framework, and used multi-headed attention mechanism. \cite{DBLP:journals/corr/abs-1712-01586} used the SRL as a sequence labeling problem, using BIO tags for labeling. Then it proposed to use the deep attentional neural network for labeling. \cite{DBLP:journals/corr/TranBM16} proposed recurrent memory network, which added memory block to RNN, and allowed us to interpret the results and discover underlying dependencies present in the data. Thus attention has been widely used in NLP. It has a big advantage that it can visualize the attention matrix to tell you which parts of the neural network are focused on when doing the task.

The design of attention gate in our RAU network is inspired by recent development of target-driven task, i.e. image classification and sentiment analysis. These tasks motivate researchers to explore an attention-oriented network structure that selectively emphasizes useful information and ignores less important information, depending on the importance of the information. For image classification, there are still some limitations and shortcomings in the application of visual attention in image classification, including hard attention, difficult to end-to-end training and so on. Aiming at these problems, this paper proposes an Recurrent Attention Unit (RAU), which can implement end-to-end training depth image classification based on attention mechanism. For sentiment analysis, the emotional words in the sentence play a key role in the emotional tendency of the whole sentence. Therefore, the weight of each word in the text can be calculated by introducing the RAU, so that the hidden state of emotional words is more significant to affective classification.
\section{Recurrent Attention Unit}

It has been proved that GRU can maintain a long period of input information \cite{DBLP:journals/corr/ChungGCB14}. However, the existing GRU architecture is difficult to adaptively pay attention to what information is exactly useful at the hidden layer state at each time step, especially when data has complex potential connections, which are common in language modeling and sentiment classification tasks. To overcome this problem, we propose a novel recurrent neutral network architecture called RAU. RAU can not only qualify which information is preserved over time but also suppress redundant information. We integrate the attention mechanism seamlessly into the interior of GRU by adding attention gates. The attention gate can enhance GRU's ability to remember long-term memory and help memory cells quickly discard unimportant content.

Since GRU and the work of this paper can be regarded as variants of RNN, we will introduce the basic structure of RAU based on GRU and explain why RAU can adaptively paying attention to key information.

\subsection{Gated Recurrent Unit}

LSTM is an improved RNN that learns long-term dependency information. It has recently been improved and promoted by \cite{Graves2014Hybrid}. GRU is a variant of LSTM that maintains the effects of LSTM while making the structure simpler, reducing training parameters and increasing the rate of model training.

\begin{figure}[h]
\centering
\includegraphics[width=2.4in, height= 1.4in]{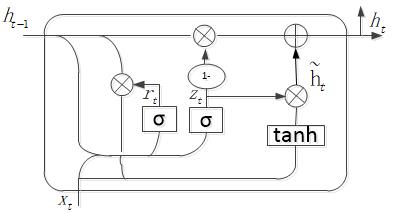}
\caption{Gated Recurrent Unit}
\label{fig:f1}
\end{figure}

 LSTM is a complex network structure that consists of three gate calculations, the forgotten gate, the input gate, and the output gate. Three gate calculations are used to increase or decrease the information in the cell state. The GRU combines the forgotten gate and the input gate in the LSTM into a single update gate, while also mixes the cell state and the hidden state. Another is a reset gate and the GRU structure is shown in  Figure~\ref{fig:f1}.  The update gate is used to control whether the information at the previous moment is brought into the current cell state.

 In Figure~\ref{fig:f1}, $\x_t $ is the input data at time $t$, $\h_t$ is the output at time $t$ of the GRU. In the GRU dynamic gate structure,  the update gate $z_t$ controls how much of the previous memory content is to be added. The update gate is computed based on the previous hidden states $\h_{t-1}$ and the current input $\x_t$:
\begin{eqnarray}\label{eq:obj1}
z_t=\sigma (\W_z[\x_t, \h_{t-1}]+\b_z),
\end{eqnarray}
where $\sigma$ represents the sigmoid function, $\W_z$ represents the update gate weight, and $\b_z$ indicates the update gate bias.

 And $r_t$ is a reset gate. After reading $\h_{t-1}$ and $\x_t$, the gate outputs a value ranging from 0 to 1, and $r_t$ indicates the percentage of information to be discarded. Value 0 means completely discarding,  allowing it to forget the previously computed state, and 1 means full reservation.
\begin{eqnarray}\label{eq:obj2}
r_t = \sigma (\W_r[\x_t,\h_{t-1}]+\b_r),
\end{eqnarray}
where $\sigma$ represents the sigmoid function, $\W_r$ represents the reset gate weight, $\b_r$ indicates the reset gate bias.

Then, a new candidate value vector will be created and added to the state. The formulas for this process are as follows:
\begin{eqnarray}\label{eq:obj3}
\tilde{\h_t} = tanh(\W_{\tilde{\h}}[x_t, r_th_{t-1}]+\b_{\tilde{\h}}),
\end{eqnarray}
where $\tanh$ is the hyperbolic tangent function, $\W_{\tilde{\h}}$ represents the update candidate value, $\b_{\tilde{\h}}$ refers to the update candidate bias, and $\tilde{\h_t}$ refers to the candidate value.

Finally, determining the value of the output, this process will retain the information of the current unit and pass it to the next unit.
\begin{eqnarray}\label{eq:obj4}
\h_t=(1-z_t)\h_{t-1}+z_t\tilde{\h_t}.
\end{eqnarray}
\subsection{Recurrent Attention Unit}

The attention mechanism is one of the important components of RAU, which is good at highlighting the important part. Whether image data or sequence data, as long as attention mechanism is in good use, it can help us find the part we need, and ignore a lot of background information. Just as people are observing images, they don't actually see every pixel of the entire image at a time. Most of them focus on specific parts of the image according to their needs. Moreover, humans can learn from the previously observed images to see where the attention of the image should be concentrated in the future. Our attention mechanism embedded in the RAU is to use attention gate to select important information to handle, and eliminate a large amount of background information that is not needed for the task at hand.
%

\begin{figure}[h]
\centering
\includegraphics[width=2.5in, height= 1.5in]{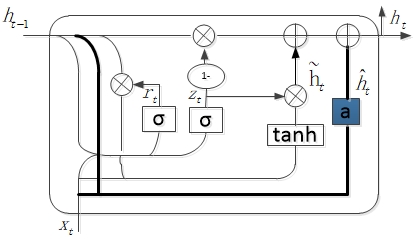} \vskip -0.2cm
\caption{Recurrent Attention Unit}
\label{fig:f2}
\end{figure}

Our attention gate receives the sequence information $x_t^1, x_t^2, \dots, x_t^n$ and hidden state at the last moment $h_{t-1}$ to learn a weight matrix $\U_t$ that can express the importance of these information.
\begin{eqnarray}\label{eq:obj5}
\bm{\alpha_t} = a(\h_{t-1},\x_t)
\end{eqnarray}
\begin{eqnarray}\label{eq:obj6}
\U_t =\frac{ \exp(\bm{\alpha_t})}{\sum_{t=1}^T \exp(\bm{\alpha_t})}
\end{eqnarray}
\begin{eqnarray}\label{eq:obj7}
\hat{\h_t}=\tanh (\U_t [\x_t, \h_{t-1}])
\end{eqnarray}
where $\x_t$ is the external m-dimensional input vector at time $t$, $\h_{t-1}$ is the n-dimensional hidden state at last moment, and $a$ is a learning function, it is only determined by $\h_{t-1}$ and $\x_t$.

In the above formula, the attention mechanism can be consider to construct a fixed length of the embedded value $\hat{\h_t}$ of the input sequence by calculating an adaptive weighted average of the state sequences $\h_{t-1}$ and input $\x_t$.

Figure~\ref{fig:f2} is a complete RAU in which the vector in the hidden state sequence $\h_{t-1}$ and  input $\x_t$ are fed into the learning function  $a(\h_{t-1},\x_t)$, resulting in a probability vector $\bm{\alpha_t}$. The vector $\hat{\h_t}$ is weighted average of $\h_{t-1}$ and  input $\x_t$, and the weight is $\U_t$.

We have reserved the update gate and reset gate in the GRU structure, and added the attention gate. Hence, we keep the values of Equation \eqref{eq:obj1} \eqref{eq:obj2}\eqref{eq:obj3} fixed and change the value of the hidden layer state of the output is as follows,
\begin{eqnarray}\label{eq:obj8}
\h_t=(1-z_t)\h_{t-1}+z_t\tilde{\h_t}/2+z_t\hat{\h_t}/2.
\end{eqnarray}

The advantage of the attention mechanism model is that the ability to integrate information over time. Therefore, by using this simplified attention mechanism, the model can help memory cells quickly discard unimportant content and increase attention to important information, thus enhancing the network's ability to remember long-term memory.

\section{Experiments}

We evaluate the performance of the RAU on four different dataset: MNIST, Fashion-MNIST, PTB and IMDB. In other words, we verified the effectiveness of RAU in image classification, language modeling and sentiment classification tasks. And these experiments show that our RAU performs better than GRU and LSTM.

\begin{table*}[t]
\centering
\begin{center}
\begin{tabular}{*{7}{c}}
\bottomrule
\multirow{2}*{} & \multicolumn{2}{c}{RAU} & \multicolumn{2}{c}{GRU}&\multicolumn{2}{c}{LSTM}\\
\cmidrule(lr){2-3}\cmidrule(lr){4-5}\cmidrule(lr){6-7}& Validation & Test &  Validation & Test  &  Validation& Test \\
\midrule
MNIST Acc. (\%) & - &98.80  & - & 98.54 & -  & 98.55\\
\midrule
Fashion-MNIST Acc. (\%) & - &89.60  & - & 87.89 & -  & 88.45\\
\midrule
IMDB Acc. (\%) & 80.42 &79.52  &79.76  &79.08  & 78.97  & 78.96\\
\midrule
\midrule
PTB-Small (perplexity) & 118.99 &113.89  &119.37  &115.06  & 120.64  & 114.96\\
\midrule
PTB-Medium (perplexity) & 86.75 &82.80  &86.50  &83.01  & 87.80  & 83.62\\
\midrule
PTB-Large (perplexity) & 82.63 &78.32  &83.10  &78.50  & 82.62  & 78.63\\

\bottomrule
\end{tabular}
\caption{Performance of three model on different tasks (perplexity is a measure of the quality of the language model, the lower the better.)}\label{table1}
\end{center}
\end{table*}

\subsection{Application to row-wise sequences of the MNIST data set}

\begin{figure}[htp]
\centering
\includegraphics[width=2in, height= 2in]{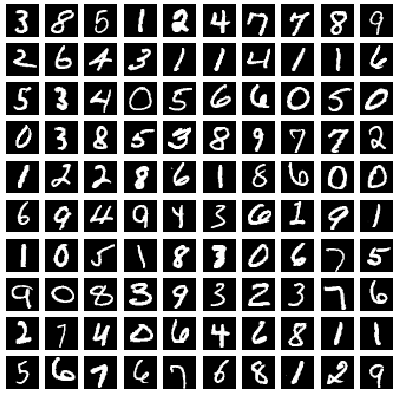}
\caption{MNIST data set examples.}
\label{fig:f3}
\end{figure}

The MNIST data set is one of the most famous data sets in machine learning, derived from the National Institute of Standards and Technology. It consists of 0-9 handwritten numeral labels, which are divided into training set and test set. Figure~\ref{fig:f3} shows a sample diagram of MNIST data for 10$\times$10.
The training set contains 60,000 handwritten digital images, while the test set contains 10,000 handwritten digital images, each of which has been normalized to a digitally centric 28$\times$28 gray-scale map. We evaluated RAU against the original GRU and LSTM on the MNIST data set  by generating the sequential input in row-wise (one row at a time). The row-wise sequence generated from each image are 28-element signal of length 28.

 In our experiments, the networks read one row at a time in scanline order (i.e. starting at the first row of the image, and ending at the last row of the image). The networks are asked to predict the category of the MNIST image only after seeing all 28 row. This is a medium long rang dependency problem because each recurrent network has 28 time steps. All networks have 128 recurrent hidden units. We stop the optimization after it converges or when it reaches 100000 iterations and report the results in Table 1 (related characteristics are listed in Table 2). It is clear from Table 1 that the proposed RAU architecture outperforms the other baseline architectures that we have tried when used same hyperparameters  such as LSTM and GRU.

\begin{table}[tb]
\begin{center}
\scalebox{0.75}[0.75]{
\begin{tabular}{|c|c|c|c|c|}
\hline
Model&\textbf{MNIST}&\makecell[c]{\textbf{Fashion} \\\textbf{MNIST} } &\textbf{IMDB}\\
\hline
\makecell{Hidden \\ Units}&28&28&128\\
\hline
\makecell{Gate\\ Activation}&Sigmoid&Sigmoid&Sigmoid\\
\hline
Activation&Softmax&Softmax&Softmax\\
\hline
Epochs&213&213&100\\
\hline
Optimizer&Adam&Adam&Adam\\
\hline
Dropout&-&-&50\%\\
\hline
Batch Size&128&128&128\\
\hline
\end{tabular}}
\caption{Network model characteristics (we have set the number of iterations to 100,000 steps, so we calculated the epoch number to be 213).}\label{table2}
\end{center}
\vskip -0.1cm
\end{table}

We also compare our RAU with recurrent model of visual attention (RAM) \cite{DBLP:journals/corr/MnihHGK14} on MNIST. RAM is a visual attention model that is formulated as a single recurrent neural network. The results are shown in Table 3. our RAU outperforms all the baseline methods on MNIST data sets. Note that RAU achieves  1.20\% test error on MNIST compared with 1.29\% test error.
\begin{table}[h]
\vskip -0.1cm
\centering

\begin{center}
\begin{tabular}{*{2}{c}}
\bottomrule
Model & Error\\
\midrule
FC, 2layers  \cite{DBLP:journals/corr/MnihHGK14} & 1.35\% \\

1 Random Glimpse \cite{DBLP:journals/corr/MnihHGK14} & 42.85\% \\

RAM, 2glimpses \cite{DBLP:journals/corr/MnihHGK14}  & 6.27\% \\

RAM, 3glimpses \cite{DBLP:journals/corr/MnihHGK14} & 2.7\% \\

RAM, 4glimpses\cite{DBLP:journals/corr/MnihHGK14}& 1.73\% \\
RAM, 5glimpses \cite{DBLP:journals/corr/MnihHGK14}& 1.55\% \\
RAM, 6glimpses \cite{DBLP:journals/corr/MnihHGK14}& \textbf{1.29}\% \\
RAM, 7glimpses \cite{DBLP:journals/corr/MnihHGK14}& 1.47\% \\
\midrule
Our model RAU&  \textbf{1.20}\% \\
\bottomrule
\end{tabular}
\caption{Comparison between traditional attention models and RAU on sequential data learning. FC denotes a fully connected network with two layers of rectifier units. Instances of the attention model are labeled with the number of glimpses.}\label{table3}
\end{center}
\end{table}

\subsection{Application to row-wise sequences of the Fashion-MNIST data set}

\begin{figure}[htb]
\vspace{-0.6cm}
\centering
\includegraphics[width=2.3in, height= 2.3in]{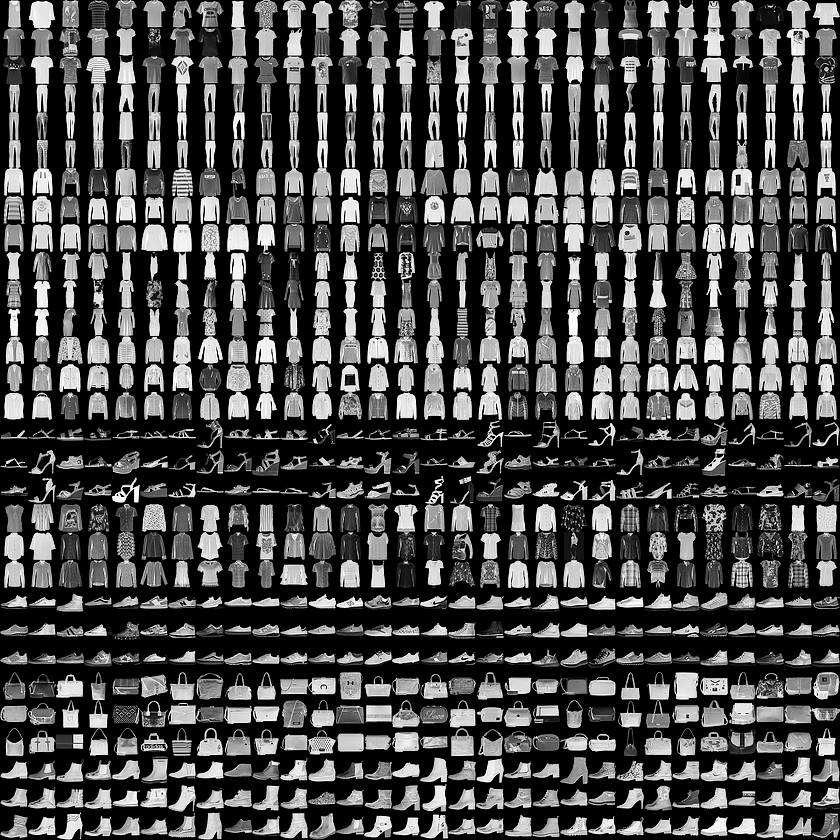} \vskip -0.1cm
\caption{Fashion-MNIST dataset example.}
\label{fig:f5}
\end{figure}

Fashion-MNIST is a data set similar to MNIST, and all items are based on the classification on the Zalando website. Each of Zalando's fashion items has a set of photographs taken by professional photographers, showing different aspects of the product, such as the appearance of the front and back, details. Fashion-MNIST also stored in 28$\times$28 gray-scale map. Figure~\ref{fig:f5} shows some examples. It contains 70000 pictures, 10 classes, and 7000 pictures per class. Randomly selected 6000 pictures from each class to form training sets, and the other 1000 to form test sets.

\begin{figure}[htb]
\vskip -0.2cm
\centering
\includegraphics[width=2.9in, height= 2.4in]{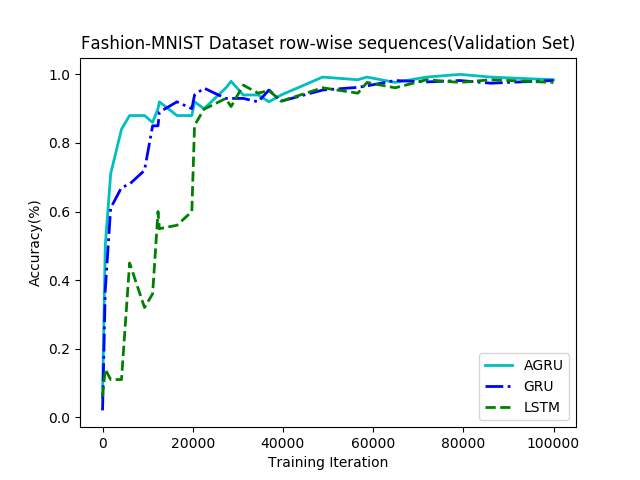} \vskip -0.2cm
\caption{Validation learning curves of three different architectures: RAU, GRU, LSTM with the same number of hidden units. The curves represent training up to 100000 steps.}
\label{fig:f6}
\vskip -0.2cm
\end{figure}

We also performed experiments on the Fashion-MNIST data set. As shown in Table 2, we compare the three architecture with same parameter configurations. We use the RAU, GRU, and LSTM network with single layer of 128 hidden units to solve the classification task. Each time we read one row of a image, it is equivalent to a image being a sequence of 28 lengths. In Figure~\ref{fig:f6}, we plotted the learning curves of three model against training steps. We observed that RAU tended to make faster progress overtime. This behavior is observed both when the number of parameters is constrained and when the number of hidden units is constrained. In Table 1, we can see with the same characteristics, our proposed RAU achieve 89.60\% accuracy in the classification task of the Fashion-MNIST data set, which is 1.71\% higher than GRU and higher than LSTM 1.15\%.

\subsection{Application to Language modeling task}
In this paper, we carried out a series of experiments on the Penn Treebank Corpus to verify that our proposed model can solve the language modeling task. The Penn Treebank Corpus \cite{Marcus:1993:BLA:972470.972475} is an English tree library, established by the University of Pennsylvania, commonly used to conduct natural language processing tasks such as part-of-speech tagging and syntactic analysis, in language mode field is also widely used.

A common indicator of the quality of a language model is perplexity. In simple terms, the perplexity value characterizes the probability of occurrence of an estimated sentence through a certain language model. For example, when you already know ($\omega_1$,$\omega_2$,$\omega_3$,$\cdots$,$\omega_m$) that the sentence appears in the corpus, then the probability of this sentence would be calculated by the language model the higher the better, which is the perplexity value as small as possible. The formula for calculating the perplexity  value is as follows:\\
\begin{equation}
\begin{aligned}
perplexity(S) =& p (\omega_1,\omega_2,\omega_3,\cdots,\omega_m)^{\frac{1}{m}}\\
 =&\sqrt[m]{\frac{1}{p (\omega_1,\omega_2,\omega_3,\cdots,\omega_m)}}\\
= &\sqrt[m]{\prod_{i=1}^m\frac{1}{p (\omega_1|\omega_1,\omega_2,\omega_3,\cdots,\omega_{i-1})}}\\
\end{aligned}
\end{equation}

The concept represented by perplexity is actually the average branch factor, which is the average number of choices when the model predicts the next word. For example, if the perplexity of a language model is 83, it means that on average, when the model predicts the next word, 83 words may be a reasonable choice for the next word.
\begin{table}[h]
\begin{center}
\scalebox{0.75}[0.75]{
\begin{tabular}{|c|c|c|c|c|}
\hline
Model&\textbf{PTB-Small}&\makecell[c]{\textbf{PTB-Mediun} } &\textbf{PTB-large}\\
\hline
Weight Initial Scale &0.1&0.05&0.04\\
\hline
Number Layer&2&2&2\\
\hline
\makecell{Hidden Units}&200&650&1500\\
\hline
\makecell{Batch Size}&20&20&20\\
\hline
Learning Rate&1&1&1\\
\hline
Learning Rate Decay&0.5&0.8&1/1.5\\
\hline
Epochs&13&35&55\\
\hline

Dropout&0&50\%&65\%\\
\hline
Vocabulary &10000&10000&10000\\
\hline
\end{tabular}}
\caption{Parameters of different sized models.}\label{table2}
\end{center}
\vskip -0.2cm
\end{table}

We implemented the three models with three different sizes of configurations in Penn Treebank Corpus. We used the same experimental setup during the experiment set (the same training, development and test data, and the same vocabulary limit). This also helps to compare different language model techniques. We first performed the experiment of PTB small size: 2 layers with 200 units in each layer. The experimental results are shown in Table 1. In the last epoch, our proposed RAU model is reachable 45.72 perplexity on the training set. And the verification set and the test set can reach the perplexity of 118.99 and 113.89 respectively. Compared with the experiments of GRU and LSTM, it is clear that the model we proposed is more effective.

 As shown in Table 4,  in the PTB-medium experiment, we reduced the initial scale of the network weight value because we want the initial value of the weight not to be too large, the smaller is beneficial to moderate training, the learning rate is unchanged, and the number of layers of the network stacked is unchanged. Here, the number of expansion steps of the gradient back propagation is increased from 20 to 30, and the number of hidden layer units is also increased by 3 times. Because the number of iterations of learning increases, the decay rate of learning rate will be smaller. As the Penn Treebank is a relatively small data set,we  prevent overfitting by dropout technology.

  In the case of a medium configuration, the loss of the three models is shown in Figure~\ref{fig:f7}. It can be seen that the proposed model can significantly reduce the perplexity of the language model.

\begin{figure}[h]
\vskip -0.2cm
\centering
\includegraphics[width=3in, height= 2.5in]{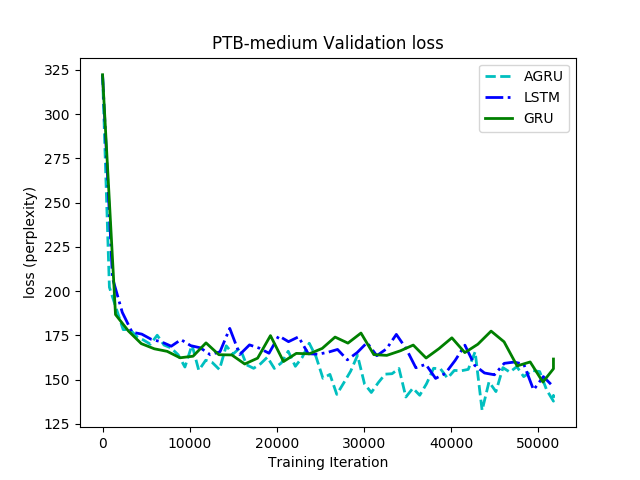} \vskip -0.2cm
\caption{The validation loss curve of the three network models (RAU, GRU and LSTM) in a medium configuration.}
\label{fig:f7}
\end{figure}

The large configuration model further reduces the initial scale of the weight values in the network, while increasing the hidden layer size to 1500, because the complexity of the model increases, the dropout rate increases from 50\% to 65\%. Comparing our results on the RAU to the results of LSTM and GRU in Table 1, we see that the RAU is highly competitive. We observed that the performance of recurrent methods depends on the size of the hidden states: they perform better as the size of the hidden states gets larger.

\subsection{Application to sentiment classification task}

We evaluate the RAU architecture on a popular document-level sentiment classification benchmark, the IMDB movie review data set \cite{maas-EtAl:2011:ACL-HLT2011}. The data set consists of a balanced sample of 25,000 positive and 25,000 negative reviews, which are divided into equal-size train and test sets, with an average document length of 231 words \cite{sentiment}. We have trained the data set on three different architectures using the two constant base learning rates of $10^{-3}$,  $10^{-4}$ and  $10^{-5}$ over 100 epochs.

\begin{figure}[h]
\vskip -0.2cm
\centering
\includegraphics[width=3in, height= 2.5in]{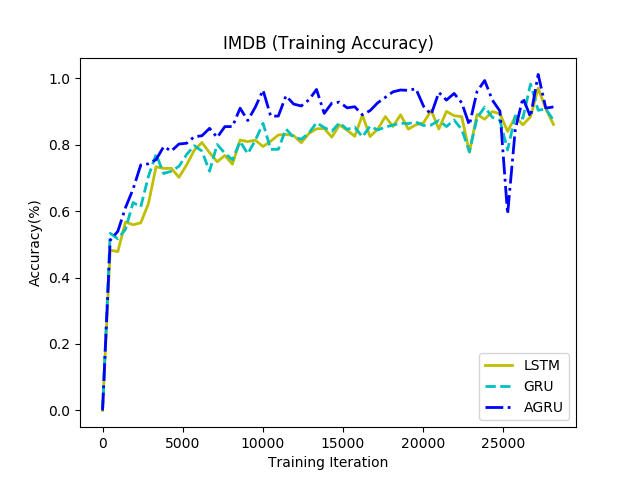} \vskip -0.2cm
\caption{Train learning curves of three different architectures: RAU, GRU, LSTM with the same  learning rate ($10^{-5}$). The curves represent training up to 30000 steps.}
\label{fig:f7}
\end{figure}
In the training, we employed 100-dimensional architecture and adopted a batch size of 32. We have observed that, using the constant base learning rate of $10^{-5}$, performance is well, whereas performance is uniformly progressing over profile-curves as shown in  Figure~\ref{fig:f7}. Table 1 summarizes the results of accuracy performance which show comparable performance among RAU, GRU, and LSTM. Table 2 also lists the number of parameters in each dataset.


\section{Conclusion}

  Attention mechanism is an intuitive method, which is widely used in the field of computer vision by giving different weights to different parts of the input. We have proposed Recurrent Attention Unit, a novel architecture that can replace GRU and LSTM in many application scenarios. It is not only  suitable for computer vision area, but also for all kind of sequence related problems. The model we proposed adds the attention gate to GRU, which not only amplifies the power of GRU but also enables the network to adaptively pay attention to the part that should be emphasized. We applied our model to three tasks: image classification, language modeling and sentiment classification. The experimental results demonstrate that our RAU outperforms both the LSTM and GRU.

\bibliographystyle{aaai}
\bibliography{mybib}

\end{document}